\pdfoutput=1

\documentclass[11pt]{article}

\usepackage{acl}

\usepackage{times}
\usepackage{latexsym}
\usepackage{amssymb}
\usepackage{multirow}
\usepackage{xcolor}
\usepackage{graphicx}

\usepackage[T1]{fontenc}

\usepackage[utf8]{inputenc}

\usepackage{microtype}
\usepackage{verbatim}
\usepackage{booktabs}

\title{A Baseline Readability Model for Cebuano}

\author{Lloyd Lois Antonie Reyes, Michael Antonio Ibañez, Ranz Sapinit, \\{\bf Mohammed Hussien, Joseph Marvin Imperial}\\
  National University \\
  Manila, Philippines \\
  \texttt{jrimperial@national-u.edu.ph} \\}

\begin{document}
\maketitle
\begin{abstract}
In this study, we developed the first baseline readability model for the Cebuano language. Cebuano is the second most-used native language in the Philippines with about 27.5 million speakers. As the baseline, we extracted traditional or surface-based features, syllable patterns based from Cebuano's documented orthography, and neural embeddings from the multilingual BERT model. Results show that the use of the first two handcrafted linguistic features obtained the best performance trained on an optimized Random Forest model with approximately 87\% across all metrics. The feature sets and algorithm used also is similar to previous results in readability assessment for the Filipino language—showing potential of crosslingual application. To encourage more work for readability assessment in Philippine languages such as Cebuano, we open-sourced both code and data\footnote{The resources can be found using this link: \url{https://github.com/imperialite/cebuano-readability}.}.


\end{abstract}

\section{Introduction}

The proper identification of the difficulty levels of reading materials is a vital aspect of the language learning process. It enables teachers and educators alike to assign appropriate materials to young learners in which they can fully comprehend, preventing boredom and disinterest ~\cite{guevarra2011development}. However, assessing readability presents challenges, particularly when you have a large corpus of text to sift through. Manually extracting and calculating a wide range of linguistic features can be time-consuming and expensive and can lead to subjectivity of labels due to human errors  \cite{deutsch2020}. To tackle this problem, more and more research in the field have focused on experimenting with automated methods for extracting possible linguistic predictors to train models for readability assessment. 

While automating readability assessment is a challenge itself, one of the original problem in the field starts with data. In the Philippines, the Mother-Tongue Based Multilingual Education (MTB-MLE) scheme was introduced by the Department of Education (DepEd) in 2013. With this initiative, there were little to no available tool for automatically assessing readability of reading resources, instructional materials, and grammatical materials in mother tongue languages aside from Filipino such as Cebuano, Hiligaynon, and Bikol ~\cite{medilo2016experience}. To answer this challenge, in this paper, we investigate various linguistic features ranging from traditional or surface-based predictors, orthography-based features from syllable patterns, and neural representations to develop a baseline readability assessment model for the Cebuano language. We use  an array of traditional machine learning algorithms to train the assessment models with hyperparameter optimization. Our results show that using non-neural features are enough to produce a competitive model for identifying the readability levels of children's books in Cebuano. 

\section{Previous Work}
Readability assessment has been the subject of research of linguistic experts and book publishers as a method of measuring comprehensibility of a given text or document. \citet{villamin1979pilipino} pioneered a readability assessment for the Filipino language in 1979. Hand-crafted indices and surface information from texts, such as hand counts of words, phrases, and sentences, are used in these formula-based techniques. An equivalent technique of traditional formula was applied on to Waray language ~\cite{oyzon2015validation} to complement the DepEd's MTB-MLE program in certain regions of the Philippines such as in Samar and Leyte. While traditional featured formulas relied on linear models, recent studies on readability research assessment have shifted their focus on expanding the traditional method to more fine-grained features. ~\citet{guevarra2011development} and \citet{macahilig2014content} introduced the use of a logistic regression model trained with unique word counts, total word and sentence counts, and mean log of word frequency. A few years later, lexical, syllable patterns, morphology, and syntactic features were eventually explored for readability of Filipino text by works of Imperial and Ong \cite{imperial2021application, imperial2020exploring, imperial2021diverse}.

\section{The Cebuano Language}
\begin{figure}[!htbp]
    \centering
    \includegraphics[width=.46\textwidth,trim={5cm 0 3cm 0}, clip]{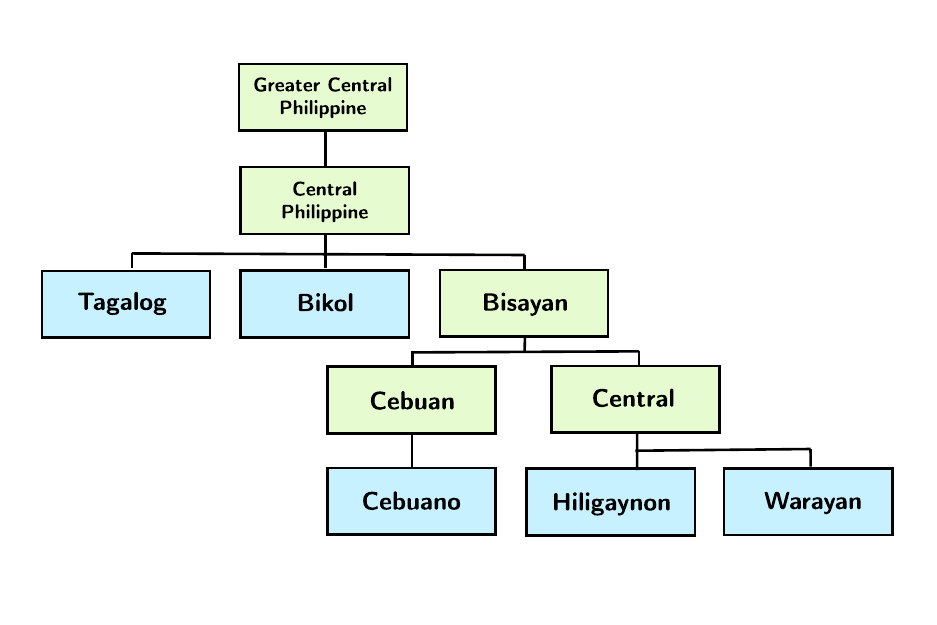}
    \caption{Right portion of the Philippine language family tree highlighting origin of Cebuano.}
    \label{language_tree}
\end{figure}

Cebuano (\textsc{ceb}) is an Austronesian language mostly spoken in the southern parts of the Philippines such as in major regions of Visayas and Mindanao. It is the language with the second highest speaker count\footnote{\url{https://www.ethnologue.com/language/ceb}} in the country with 27.5 million, just after Tagalog, where the national language is derived from, with 82 million speakers. Both Cebuano and Tagalog languages observe linguistic similarities such as in derivation, prefixing, disyllabic roots, and reduplication \cite{blake1904differences}. On the other hand, differences are seen in syntax such as use of particles (\textit{ay}, \textit{y}), phonetic changes, and morphological changes on verbs. Figure~\ref{language_tree} illustrates a portion of the Philippine language family tree emphasizing on where Cebuano originated. Cebuano is part of the Central Philippine subtree along with Tagalog and Bikol which can be attributed to their similarities and differences as mentioned. The full image can be viewed at \citet{oco2013dice}.

\subsection{Cebuano Readability Corpus}
We compiled the first Cebuano text corpus composed of 277 expert-annotated literary pieces uniform to the first three grade levels (L1, L2, and L3) of the Philippine primary education. For comparison to international grading systems, the standard age range for each level is 6-7, 7-8, and 8-9 respectively. We collected the materials from three online, open-sourced book repositories online: \textbf{Let's Read}, \textbf{Bloom Library}, and \textbf{DepEd Commons}. All materials are licensed under Creative Commons BY 4.0 allows redistribution in any medium or format provided proper attribution. Table~\ref{data} shows the distribution of the collected corpus. \\

\begin{table}[!htbp]
\small
\centering
\begin{tabular}{lllll}
\toprule
\textbf{Corpus} & \textbf{L1} & \textbf{L2} & \textbf{L3} & \textbf{Total} \\
\midrule
Let's Read & 6 & 21 & 50 &  82 \\
Bloom & 50 & 50 & 25 &  125 \\
DepEd & 22 & 1 & 4 &  27 \\
\midrule
 \bf Total &  \bf 76 &  \bf 72 &  \bf 79 &  \bf 227 \\
\bottomrule
\end{tabular}
\caption{Distribution of compiled text passages in Cebuano.}
\label{data}
\end{table}

\noindent\textbf{Let’s Read.} Let’s Read\footnote{\url{https://www.letsreadasia.org/}} is an initiative by the The Asia Foundation to open-source culturally friendly children's books in diverse themes, characters, and settings. The resource materials from this repository are mostly sourced from BookLabs and translated by local volunteers across multiple languages including Cebuano. Let's Read covers a wide variety of genre such as gender equality, environment, understanding and empathy, and science and technology. We collected 82 Cebuano children's books from this website for our corpus. \\

\noindent\textbf{Bloom.} The Bloom Library\footnote{\url{https://bloomlibrary.org/}} is also an free repository of diverse children's books resources funded and maintained by the Summer Institute of Linguistics (SIL International). Similar to Let's Read, local volunteers can also upload high-quality and validated translations of book resources or original pieces to the platform. We collected 125 Cebuano children's books from this website for our corpus.\\

\noindent\textbf{DepEd Commons.} The Commons Library\footnote{\url{https://commons.deped.gov.ph/}} is an initiative by the Department of Education in the Philippines to grant free access to literature in various Philippine languages for students and teachers during the COVID-19 pandemic. We collected 27 Cebuano children's books from this website for our corpus.\\

\section{Linguistic Features}
In this study, we extracted three linguistic feature groups from our Cebuano text corpus: \textbf{traditional or surface-based features}, \textbf{orthography-based features}, and \textbf{neural embeddings}. To the best of our knowledge, no study has ever been conducted to assess and explore the readability assessment of Cebuano text using these features. 

\subsection{Traditional Features (TRAD)}
Traditional or surface-based features are predictors that were used by experts for their old readability formulas for Filipino such as sentence and word counts in \citet{guevarra2011development}. Despite the claims that these features insufficiently measures deeper text properties for readability assessment \cite{redish2000readability}, since this is the pioneering study for Cebuano, we still considered these features for our baseline model development. In this study, we adapted the seven features of traditional features from existing works in Filipino ~\cite{imperial2020exploring, imperial2021application, imperial2021diverse} such as \textit{number of unique words, number of words, average word length, average number of syllables, total number of sentences, average sentence length} and \textit{number of polysyllable words}. 

\subsection{Syllable Pattern (SYLL)}
Orthography-based features measure character-level complexity of texts through combinations of various syllable patterns ~\cite{imperial2021diverse}. Same as in Filipino, we adapted syllable patterns as features for the baseline model development but used only seven recognizable consonant-vowel combinations linguistically documented in the Cebuano language \cite{blake1904differences}. We used \textit{consonant clusters} and syllable pattern combinations of \textit{v, cv, cc, vc, cvc, ccv, ccvc} normalized by the number of words.

\subsection{Substitute Features using Neural Embeddings (NEURAL)}
The use of Transformer-based language model embeddings have shown to be an effective \textit{substitute} for handcrafted features in low-resource languages \cite{imperial-2021-bert}. Probing tasks have shown that these representations contain information such as semantic and syntactic knowledge \cite{rogers-etal-2020-primer} which can be useful in readability assessment. For this study, we extracted embedding representations with dimension size of 768 from the multilingual BERT model \cite{devlin-etal-2019-bert} as features for each instance from the Cebuano corpus. According to the training recipe of multilingual BERT, Cebuano data in the form of Wikipedia dumps was included in its development which makes the model a viable option for this study.



\begin{table}[!htbp]
\small
\centering
\begin{tabular}{ccccc}
\toprule
\textbf{Feature} & \textbf{Acc} & \textbf{Prec} & \textbf{Rec} & \textbf{F1} \\
\midrule

\bf TRAD & \bf 0.789 & \bf 0.754 & \bf 0.749 & \bf 0.750 \\
SYLL & 0.544 & 0.546 & 0.559 & 0.551 \\

\midrule
TRAD + SYLL & 0.719 & 0.721 & 0.722 & 0.718 \\
NEURAL & 0.754 & 0.759 & 0.766 & 0.757 \\
Combination & 0.737 & 0.714 & 0.729 & 0.714 \\
\bottomrule
\end{tabular}
\caption{Performance of finetuned Logistic Regression model.}
\label{table1}
\end{table}

\begin{table}[!htbp]
\small
\centering
\begin{tabular}{ccccc}
\toprule
\textbf{Feature} & \textbf{Acc} & \textbf{Prec} & \textbf{Rec} & \textbf{F1} \\
\midrule

TRAD & 0.718 & 0.728 & 0.685 & 0.676 \\
SYLL & 0.649 & 0.648 & 0.648 & 0.646 \\

\midrule
TRAD + SYLL & 0.789 & 0.787 & 0.791 & 0.784 \\
\textbf{NEURAL} & \textbf{0.807} & \textbf{0.813} & \textbf{0.812} & \textbf{0.811} \\
Combination & 0.789 & 0.788 & 0.789 & 0.793 \\
\bottomrule
\end{tabular}
\caption{Performance of finetuned Support Vector Machines model.}
\label{table2}
\end{table}

\begin{table}[!htbp]
\small
\centering
\begin{tabular}{ccccc}
\toprule
\textbf{Feature} & \textbf{Acc} & \textbf{Prec} & \textbf{Rec} & \textbf{F1} \\
\midrule

TRAD & 0.842 & 0.843 & 0.842 & 0.842 \\
SYLL & 0.579 & 0.579 & 0.586 & 0.580 \\

\midrule
\textbf{TRAD + SYLL} & \textbf{0.873} & \textbf{0.852}  & \textbf{0.858} & \textbf{0.852} \\
NEURAL & 0.772 & 0.776 & 0.761 & 0.763 \\
Combination & 0.825 & 0.801 & 0.804 & 0.799 \\
\bottomrule
\end{tabular}
\caption{Performance of finetuned Random Forest model.}
\label{table3}
\end{table}

\section{Experiment Setup}
The task at hand is a multiclass classification problem with three classes being the aforementioned grade levels. We specifically chose traditional learning algorithms such as Logistic Regression, Support Vector Machines, and Random Forest for building the baseline models for post-training interpretation techniques described in the succeeding sections. To reduce bias, a $k$-fold cross validation where $k=5$ was implemented. For the intrinsic evaluation, we used standard metrics such as accuracy, precision, recall and macro F1-score. In addition, we also used grid search to optimize the following model-specific hyperparamters: solver and regularization penalties for Logistic Regression, kernel type, maximum iterations, and regularization penalties for Support Vector Machines, and number of estimators, maximum features, and maximum depth for Random Forest.

\section{Results}
To assess the effectiveness of the proposed framework in the experimentation, we examined model performances on three different ablation studies: (a) linguistic features only, (b) neural embeddings only, and (c) combination of the two via concatenation. The results of each fine-tuned model utilizing the given evaluation metric are showed in Tables~\ref{table1},~\ref{table2}, and~\ref{table3}. 

Across the board, the best performing model and feature combination for Cebuano achieved approximately 87.3\% for all metrics using the combination of TRAD and SYLL features with Random Forest. This top performing model makes used of 100 tree estimators, automatically adjusted maximum features, and a max depth of 20. Interestingly, the feature combination and the algorithm of choice is also the \textit{same} for Filipino readability assessment as seen in the work of\citet{imperial2021diverse}. This may suggest that, despite language differences and similarities, the use of surface-based features such as counts and syllable patterns are accepted for both Filipino and Cebuano languages in the readability assessment task. Referring again to Figure~\ref{language_tree} for emphasis, both languages are part of the Central Philippine subtree which opens the possibility of a cross-lingual application of linguistic features for future research.

This effectiveness of surface-based features is also seen for the optimized Logistic Regression model where using TRAD features obtained the best performance. In the case of the optimized Support Vector Machine model, the use of neural embeddings alone obtained better scores than the combination of traditional and syllable pattern features. This result affirms the observation in \citet{imperial-2021-bert} where the extracted neural embeddings can serve as substitute features and can relatively be at par with handcrafted features. 

\section{Discussion}
\subsection{Model Interpretation}
\begin{figure}[h]
    \centering
    \includegraphics[width=.45\textwidth]{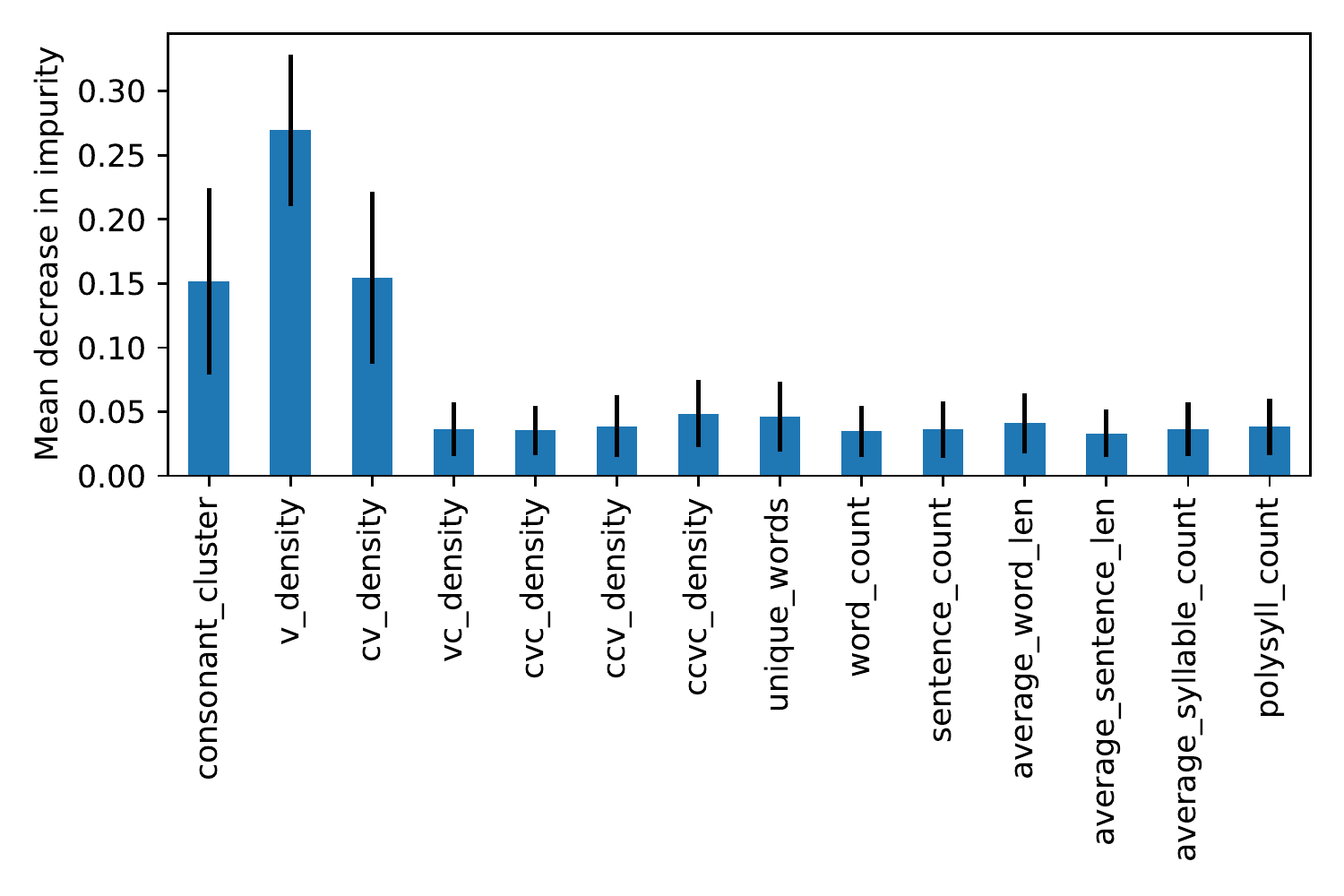}
    \caption{Feature importance by mean decrease impurity.}
    \label{feature_importance_mdi}
\end{figure}

\begin{figure}[h]
    \centering
    \includegraphics[width=.45\textwidth]{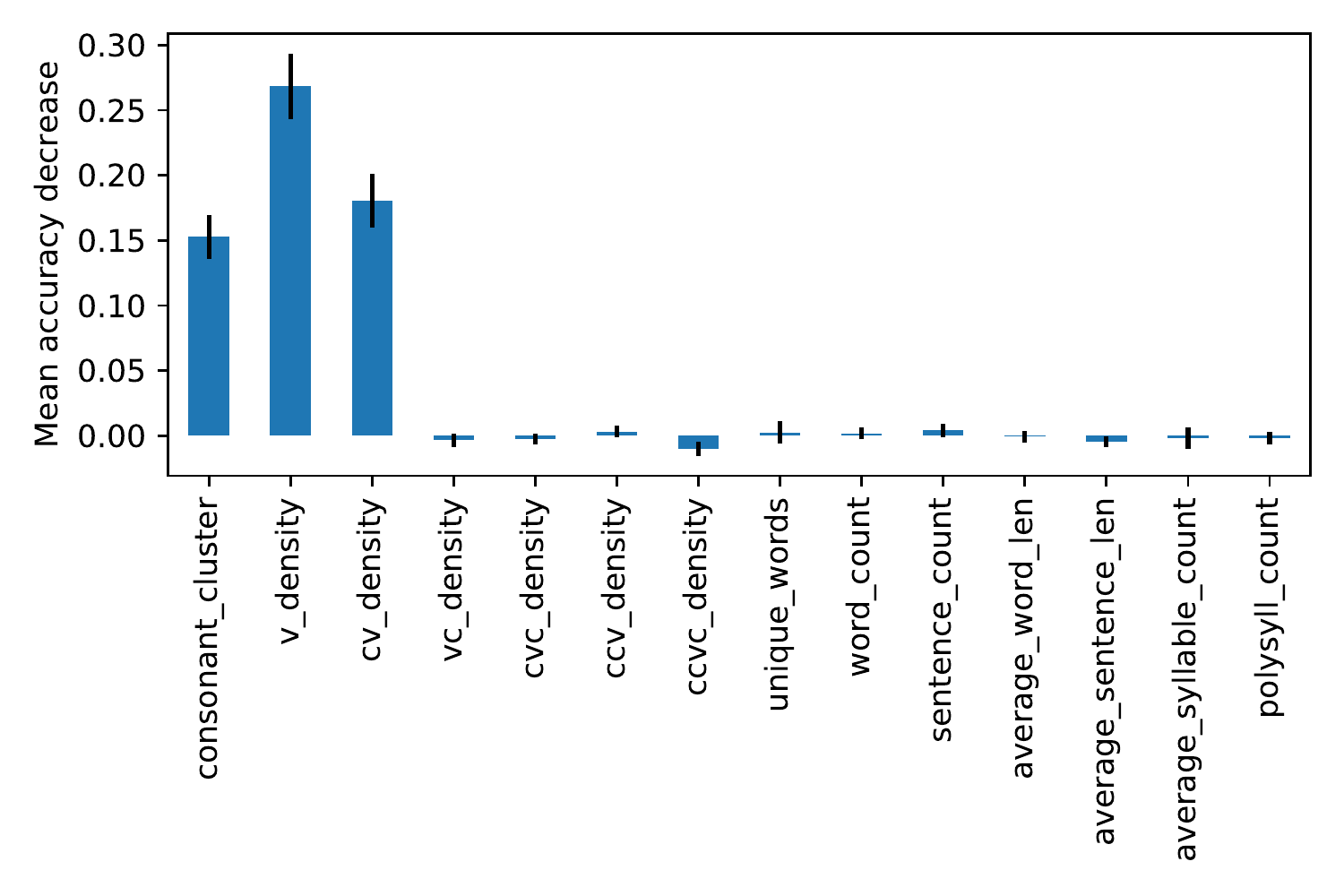}
    \caption{Feature importance by permutation on full model.}
    \label{feature_importance_full}
\end{figure}

To understand more about which specific linguistic feature is contributive during model training, we used two versions of model interpretation algorithms specifically used for Random Forest models: \textbf{permutation on full model} and \textbf{mean decrease in impurity (MDI)} as shown in Figures~\ref{feature_importance_full} and ~\ref{feature_importance_mdi} respectively. Feature permutation recursively adds a predictor to a null model and evaluates the growth in accuracy while mean decrease impurity adds up all weighted impurity score reductions or homogeneity averaged for all tree estimators \cite{breiman2001random}. From both the feature importance results, the most important feature is the \textit{v\_density} or \textit{singular vowel density}. This may indicate that the denser the vowels in a word, the more complex the text becomes. Likewise, both \textit{cv\_density} and \textit{consonant clusters} emerged as second top predictors for both analysis which may suggest that in Cebuano, words with combined consonants with no intervening vowels are more apparent in complex sentences than from easier ones. 


%
%

\subsection{Feature Correlation}
We also looked at model-independent feature analysis techniques through \textbf{Spearman correlation} with respect to readability levels. Table~\ref{featureRankingCorrelation} shows the top ten highly correlated features. In support to the findings described in Sections 6 and 7.1, all correlated linguistic features belong to the TRAD and SYLL feature sets with \textit{number of unique words} at the top. This may suggest that the density of unique words may increase relative to the readability level in a positive direction. In addition, \textit{cv}, \textit{cvc}, and \textit{ccv} densities are the only syllable pattern features that placed top in both model-dependent and independent feature interpretation techniques. This may hint further potential as readability predictors for other text domains. To note, the \textit{cv}-pattern in Cebuano is one of the most common consonant-vowel combinations \cite{zorc1976bisayan,yap2019cebuano}.

%
%

\begin{table}[!htbp]
\small
\begin{center}
\begin{tabular}{|c|c|c|}
\hline \bf Feature Set & \bf Predictor & \bf $\rho$ \\
\hline {TRAD} & unique\_words & 0.337 \\
\hline {SYLL} & cv\_density & 0.327\ \\
\hline \multirow{2}{*}{TRAD} & word\_count & 0.298 \\
 & average\_sentence\_len & 0.295 \\
\hline {SYLL} & cvc\_density & 0.293 \\
\hline {TRAD} & sentence\_count & 0.292 \\
\hline \multirow{2}{*}{SYLL} & consonant\_cluster & 0.293 \\
 & ccv\_density & 0.217 \\
\hline {TRAD} & polysyll\_count & 0.192 \\
\hline {SYLL} & vc\_density & 0.190 \\
\hline
\end{tabular}
\caption{Feature ranking using Spearman correlation. }
\label{featureRankingCorrelation}
\end{center}
\end{table}

\section{Outlook}
We developed the first ever baseline machine learning model for readability assessment in Cebuano. Among the three linguistic feature groups extracted to build the model, the combination of traditional or surface-based features (TRAD) with syllable pattern based features (SYLL) produced the highest performance using an optimized Random Forest model. One of the main challenges in the field is the limited amount of resource for tools and data especially for low-resource languages \cite{vajjala2021trends}. To answer this call and encourage growth of research in this direction, we open-sourced the compiled dataset of annotated Cebuano reading materials and the code for model development.

\section*{Acknowledgements}
The authors would like to thank the anonymous reviewers for their valuable feedback. This project is supported by the Google AI Tensorflow Faculty Grant awarded to Joseph Marvin Imperial.

\bibliography{anthology,custom}
\bibliographystyle{acl_natbib}

\end{document}